\documentclass[runningheads]{llncs}
\usepackage{graphicx}
\usepackage{comment}
\usepackage{amsmath,amssymb} % define this before the line numbering.
\usepackage{color}
\usepackage[breaklinks=true,bookmarks=false]{hyperref}
\usepackage{bm}
\usepackage[ruled]{algorithm2e}

\usepackage{booktabs}
\usepackage{threeparttable}
\usepackage{setspace}
\usepackage{multirow}
\usepackage{makecell}
\usepackage{marvosym}
\usepackage{appendix}

\begin{document}
% \linenumbers
\pagestyle{headings}
\mainmatter
\def\ECCVSubNumber{2338}  % Insert your submission number here

\title{Knowledge Transfer via Dense Cross-Layer Mutual-Distillation} % Replace with your title

\begin{comment}
\titlerunning{ECCV-20 submission ID \ECCVSubNumber}
\authorrunning{ECCV-20 submission ID \ECCVSubNumber}
\author{Anonymous ECCV submission}
\institute{Paper ID \ECCVSubNumber}
\end{comment}
%******************

%\begin{comment}
\titlerunning{Knowledge Transfer via Dense Cross-Layer Mutual-Distillation}
\author{Anbang Yao\inst{}\thanks{Equal contribution. \textsuperscript{\Letter} Corresponding author. Experiments were mostly done by Dawei Sun when he was an intern at Intel Labs China, supervised by Anbang Yao.}\textsuperscript{\Letter}
\and Dawei Sun\inst{}${^\star}$}
\authorrunning{A. Yao and D. Sun}
\institute{Intel Labs China\\
\email{\{anbang.yao,dawei.sun\}@intel.com}}
%\end{comment}
%******************
\maketitle

\begin{abstract}
Knowledge Distillation (KD) based methods adopt the one-way Knowledge Transfer (KT) scheme in which training a lower-capacity student network is guided by a pre-trained high-capacity teacher network. Recently, Deep Mutual Learning (DML) presented a two-way KT strategy, showing that the student network can be also helpful to improve the teacher network. In this paper, we propose Dense Cross-layer Mutual-distillation (DCM), an improved two-way KT method in which the teacher and student networks are trained collaboratively from scratch. To augment knowledge representation learning, well-designed auxiliary classifiers are added to certain hidden layers of both teacher and student networks. To boost KT performance, we introduce dense bidirectional KD operations between the layers appended with classifiers. After training, all auxiliary classifiers are discarded, and thus there are no extra parameters introduced to final models. We test our method on a variety of KT tasks, showing its superiorities over related methods. Code is available at \url{https://github.com/sundw2014/DCM}.

\end{abstract}

\section{Introduction}

In recent years, deep Convolutional Neural Networks (CNNs) have achieved remarkable success in many computer vision tasks such as image classification~\cite{ref01}, object detection~\cite{ref02} and semantic segmentation~\cite{ref03}. However, along with the rapid advances on CNN architecture design, top-performing models ~\cite{ref04,ref05,ref06,ref07,ref08,ref09,ref10} also pose intensive memory, compute and power costs, which limits their use in real applications, especially on resource-constrained devices.

\begin{figure}[t]
\begin{center}
\includegraphics[width=0.7\linewidth]{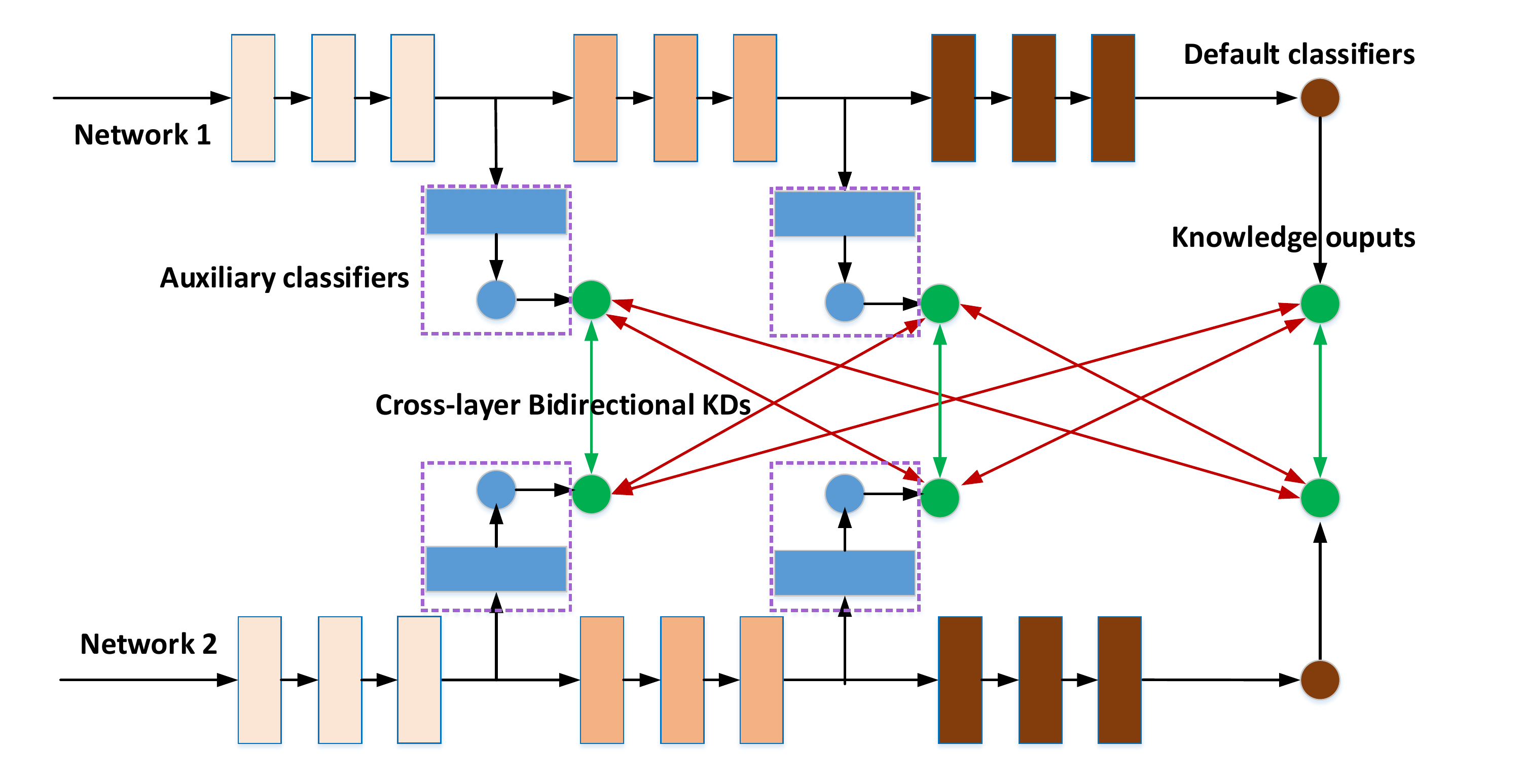}
\end{center}
\vskip -0.2 in
   \caption{Structure overview of the proposed method. For illustration, auxiliary classifiers are added to two hidden layers of each network, which are removed after training. Green/red arrows denote bidirectional knowledge distillation operations between the same-staged/different-staged layers of two networks. Best viewed in color.}
\label{fig:framework}
\vskip -0.2 in
\end{figure}

To address this dilemma, Knowledge Transfer (KT) attracts great attentions among existing research efforts~\cite{ref11}. KT is typically treated as a problem of transferring learnt information from one neural network model to another. The first attempt of using KT to cope with model compression was made in~\cite{ref12} where Bucil{\v{a}} et al. used an ensemble of neural networks trained on a small annotated dataset to label a much larger unlabelled dataset. By doing this, a smaller and faster neural network can be trained using many more labeled samples, and thus the final model can have much better performance than that is trained solely on the original small dataset.~\cite{ref13} extended this idea to train a shallower and wider neural network. Hinton et al. advanced knowledge transfer research by introducing the well-known Knowledge Distillation (KD)~\cite{ref14} method adopting a teacher-student framework. In KD, a lower-capacity student network is enforced to mimic the probabilistic outputs by a pre-trained high-capacity teacher network as well as the one-hot ground-truth labels. Naturally, the teacher can also be an ensemble of multiple models. Since then, numerous KD variants~\cite{ref15,ref16,ref17,ref18,ref19,ref58,ref50-2} have been proposed, mostly focusing on using either feature maps or attention maps at the intermediate layers of the teacher network as the extra hints for improving KD designs. Following~\cite{ref14}, these methods adopted the same teacher-student framework in which the teacher network is trained beforehand and fixed under the assumption that it always learns a better representation than the student network. Consequently, they all used the one-way KT strategy, where knowledge can only be transferred from a teacher network to a student network. Recently, Deep Mutual Learning (DML)~\cite{ref20} was proposed, which achieved superior performance by a powerful two-way KT design, showing that the probabilistic outputs from the last layer of both teacher and student networks can be beneficial to each other.

\textit{In this paper, we restrict our focus to advance two-way KT research in the perspective of promoting knowledge representation learning and transfer design}. Dense Cross-layer Mutual-distillation (DCM), an improved two-way KT method which is capable of collaboratively training the teacher and student networks from scratch, is the main contribution of this paper. Fig.~\ref{fig:framework} shows the structure overview of DCM. Following the deep supervision methodology~\cite{ref21,ref05,ref44,ref58}, we first add well-designed auxiliary classifiers to certain hidden layers of both teacher and student networks, allowing DCM to capture probabilistic predications not only from the last layer but also from the hidden layers of each network. To the best of our knowledge, deep supervision design is overlooked in the knowledge transfer field. Furthermore, we present dense cross-layer bidirectional KD operations to regularize the joint training of the teacher and student networks. On the one hand, knowledge is mutually transferred between the same-staged supervised layers. On the other hand, we find the bidirectional KD operations between the different-staged supervised layers can further improve KT performance, thanks to the well-designed auxiliary classifiers which alleviate semantic gaps of the knowledge learnt at different-staged layers of two networks. Note that there are no extra parameters added to final models as all auxiliary classifiers are discarded after training. Experiments are performed on image classification datasets with a variety of KT settings. Results show that our method outperforms related methods by noticeable margins, validating the importance of connecting knowledge representation learning with bidirectional KT design.

\section{Related Work}
In this section, we briefly summarize existing works related to our method.

\textbf{Knowledge Distillation Applications.} Although KD based methods were primarily proposed for model compression~\cite{ref14,ref15,ref16,ref17,ref18,ref19,ref50-2,ref60}, there have been many attempts to extend them to other tasks recently. Two representative examples are lifelong learning and multi-modal visual recognition. In lifelong learning task, the combination of KD and other techniques such as fine-tuning and retrospection was applied, targeting to adapt a pre-trained model to new tasks while preserving the knowledge gained on old tasks~\cite{ref22,ref23,ref52}. When designing and training multiple-stream networks dedicated to action recognition~\cite{ref24}, person re-identification~\cite{ref25}, depth estimation and scene parsing~\cite{ref26,ref27}, cross-modal distillation was used to facilitate the knowledge transfer between the network branches trained on different sources of data such as RGB and depth. Other KD extensions include but are not limited to efficient network design~\cite{ref28,ref29}, style transfer~\cite{ref31}, machine translation~\cite{ref30,ref59} and multi-task learning~\cite{ref53}. Our method differs from these approaches in task and formulation.

\textbf{Co-Training.} Blum and Mitchell proposed a pioneering co-training framework~\cite{ref32} %in the context of using both labeled and unlabeled data where the same instances are split into two distinct views.
in which two models were trained separately on each of two views of labeled data first, and then more unlabelled samples as well as the predictions by each trained model were used to enlarge training data size. Recently, several deep co-training schemes have been proposed, mostly following the semi-supervised learning paradigm.~\cite{ref33} extended the idea of~\cite{ref32} via presenting a deep adversarial co-training method that uses adversarial samples to prevent multiple neural networks trained on different views from collapsing into each other.~\cite{ref34} proposed a cooperative learning mechanism in which two agents handling the same visual recognition task can transfer their current knowledge learnt on different sources of data to each other.~\cite{ref35} addressed multi-task machine translation problem with a dual co-training mechanism.~\cite{ref55} considered the co-training of several classifier heads of the same network. Unlike these methods, we aim to improve the two-way knowledge transfer design for supervised image classification task.

\textbf{Deep Supervision.} The basic idea of deep supervision is to add extra classifiers to the hidden layers of a deep CNN architecture, which will be removed after training. It was originally proposed in~\cite{ref05,ref21} to combat convergence issue when designing and training deep CNNs for image recognition task. Since then, it has been widely used in training deep CNN architectures specially designed to handle other visual recognition tasks such as edge detection~\cite{ref36}, object detection~\cite{ref38,ref39}, semantic segmentation~\cite{ref40,ref41}, human pose estimation~\cite{ref42,ref43} and anytime recognition~\cite{ref44,ref54}. In this paper, we extend the idea of deep supervision to promote the two-way knowledge transfer research.

\section{Proposed Method}

In this section, we detail the formulation and implementation of our method.

\subsection{KD and DML}

We first review the formulations of Knowledge Distillation (KD)~\cite{ref14} and Deep Mutual Learning (DML)~\cite{ref20}. For simplicity, we follow the teacher-student framework, and only consider the very basic case where there are one single teacher network and one single student network. Given the training data $X={\{x_n\}}_{n=1}^N$ consisting of $N$ samples collected from $M$ image classes, the ground-truth labels are denoted as $Y={\{y_n\}}_{n=1}^N$. Let $W_{t}$ be a teacher network trained beforehand and fixed, and let $W_{s}$ be a student network. In KD, the student network $W_{s}$ is trained by minimizing
\begin{equation}\label{eq:01}
\begin{aligned}
L_{s}=L_{c}(W_{s}, X, Y) + \lambda L_{kd}(\hat P_{t}, \hat P_{s}), \\
\end{aligned}
\end{equation}
where $L_{c}$ is the classification loss between the predications of the student network and the one-hot ground-truth labels, $L_{kd}$ is the distillation loss, $\lambda$ is a coefficient balancing these two loss terms. In~\cite{ref14}, $L_{kd}$ is defined as
\begin{equation}\label{eq:02}
\begin{aligned}
L_{kd}(\hat P_{t}, \hat P_{s}) = -\frac{1}{N} \sum\limits_{n=1}^N\sum\limits_{m=1}^{M}\hat P_{t}^m{(x_n)}\log\hat P_{s}^m{(x_n)}. \\
\end{aligned}
\end{equation}
Given a training sample $x_n$, its probability of image class $m$ is computed as
\begin{equation}\label{eq:03}
\begin{aligned}
\hat P^m(x_n)=\frac{exp(z_n^m/T)}{\sum\nolimits_{m=1}^{M}exp(z_n^m/T)}, \\
\end{aligned}
\end{equation}
where ${z_n^m}$ is the output logit for image class $m$ obtained from the last layer (i.e., default classifier) of a neural network, and $T$ is a temperature used to soften the probabilistic outputs. The distillation loss defined by~Eq.~\ref{eq:02} can be considered as a modified cross-entropy loss using the probabilistic outputs of the pre-trained teacher network as the soft labels instead of the one-hot ground-truth labels.

Now it is clear that KD encourages the student network to match the probabilistic outputs of the pre-trained teacher model via a one-way Knowledge Transfer (KT) scheme. \textit{Two key factors to KD based methods are: the representation of knowledge and the strategy of knowledge transfer}. DML considers the latter one by presenting a two-way KT strategy in which the probabilistic outputs from both teacher and student networks can be used to guide the training of each other. DML can be viewed as a bidirectional KD method that jointly trains the teacher and student networks via interleavingly optimizing two objectives:
\begin{equation}\label{eq:04}
\begin{aligned}
L_{s}=L_{c}(W_{s}, X, Y) + \lambda L_{dml}(\hat P_{t}, \hat P_{s}) \\
L_{t}=L_{c}(W_{t}, X, Y) + \lambda L_{dml}(\hat P_{s}, \hat P_{t}). \\
\end{aligned}
\end{equation}
Here, $\lambda$ is set to 1 and fixed~\cite{ref20}. As for the definition of the distillation loss $L_{dml}$, instead of using~Eq.~\ref{eq:02}, DML uses Kullback-Leibler divergence:
\begin{equation}\label{eq:05}
\begin{aligned}
L_{dml}(\hat P_{t}, \hat P_{s}) = \frac{1}{N} \sum\limits_{n=1}^N\sum\limits_{m=1}^{M}\hat P_{t}^m{(x_n)}\log \frac{\hat P_{t}^m{(x_n)}}{\hat P_{s}^m{(x_n)}}. \\
\end{aligned}
\end{equation}
The $\hat P^m(x_n)$ is the same as that in KD. KL divergence is equivalent to cross entropy from the perspective of gradients calculation. Unlike KD containing two separate training phases, DML can jointly train the teacher and student networks in an end-to-end manner, and shows much better performance. This is attributed to the two-way KT strategy. However,
%DML only considers the probabilistic outputs from the last layer of a network as the soft labels to assist the training of the other network. That is,
the information contained in the hidden layers of networks has not been explored by DML. Moreover, the problem of connecting more effective knowledge representation learning with bidirectional KT design has also not been
studied by DML.

\subsection{Dense Cross-layer Mutual-distillation}
Our DCM promotes DML via jointly addressing two issues discussed above.

%\noindent
\textbf{Knowledge representation learning with deep supervision}. Ideally, the knowledge should contain rich and complementary information learnt by a network and can be easily understood by the other network. Recall that many KD variants~\cite{ref15,ref16,ref17,ref18,ref19,ref57} have validated that feature maps or attention maps extracted at the hidden layers of a pre-trained teacher network are beneficial to improve the training of a student network under the premise of using the one-way KT scheme. Being a two-way KT method, instead of using either intermediate feature maps or attention maps extracted in an unsupervised manner as the additional knowledge, our DCM adds relevant auxiliary classifiers to certain hidden layers of both teacher and student networks, aggregating probabilistic knowledge not only from the last layer but also from the hidden layers of each network. This is also inspired by the deep supervision methodology~\cite{ref05,ref21} which is overlooked in the knowledge transfer research. As showed in our experiments and~\cite{ref44,ref41,ref54,ref58}, even adding well-designed auxiliary classifiers to the hidden layers of a modern CNN can only bring marginal or no accuracy improvement. This motivates us to present a more elaborate bidirectional KT strategy.

%\noindent
\textbf{Cross-layer bidirectional KD}. With default and well-designed auxiliary classifiers, rich probabilistic outputs learnt at the last and hidden layers of both teacher and student networks can be aggregated on the fly during the joint training. Moreover, these probabilistic outputs are in the same semantic space, and thus our DCM introduces dense cross-layer bidirectional KD operations to promote the two-way KT process, which are illustrated in Fig.~\ref{fig:framework}.

%\noindent
\textbf{Formulation}. In the following, we detail the formulation of DCM. We follow the notations in the last sub-section. Let $Q={\{(t_k,s_k)\}}_{k=1}^K$ be a set containing $K$ pairs of the same-staged layer indices of the teacher network $W_{t}$ and the student network $W_{s}$, indicating the locations where auxiliary classifiers are added. Let $(t_{K+1},s_{K+1})$ be the last layer indices of $W_{t}$ and $W_{s}$, indicating the locations of default classifier. DCM simultaneously minimizes the following two objectives:
\begin{equation}\label{eq:07}
\begin{aligned}
L_{s}=L_{c}(W_{s}, X, Y) + \alpha L_{ds}(W_{s}, X, Y) + \beta L_{dcm_1}(\hat P_{t}, \hat P_{s}) + \gamma L_{dcm_2}(\hat P_{t}, \hat P_{s}) \\
L_{t}=L_{c}(W_{t}, X, Y) + \alpha L_{ds}(W_{t}, X, Y) + \beta L_{dcm_1}(\hat P_{s}, \hat P_{t}) + \gamma L_{dcm_2}(\hat P_{s}, \hat P_{t}), \\
\end{aligned}
\end{equation}
where $L_{s}$/$L_{t}$ is the loss of the student/teacher network. In this paper, we set $\alpha$, $\beta$, $\gamma$ and $T$ to 1 and keep them fixed owing to easy implementation and satisfied results (\textit{In fact, we tried the tedious manual tuning of these parameters, but just got marginal extra gains compared to this uniform setting}). Note that the teacher and student networks have the same loss definition. For simplicity, we take $L_{s}$ as the reference and detail its definition in the following description. In $L_{s}$, $L_{c}$ denotes the default loss which is the same as that in KD and DML. $L_{ds}$ denotes the total cross-entropy loss over all auxiliary classifiers added to the different-staged layers of the student network, which is computed as
\begin{equation}\label{eq:08}
\begin{aligned}
L_{ds}(W_{s}, X, Y) = \sum\limits_{k=1}^{K}L_{c}(W_{s_k}, X, Y). \\
\end{aligned}
\end{equation}
$L_{dcm_1}$ denotes the total loss of the same-staged bidirectional KD operations, which is defined as
\begin{equation}\label{eq:09}
\begin{aligned}
L_{dcm_1}(\hat P_{t}, \hat P_{s}) = \sum\limits_{k=1}^{K+1} L_{kd}(\hat P_{t_k}, \hat P_{s_k}). \\
\end{aligned}
\end{equation}
$L_{dcm_2}$ denotes the total loss of the different-staged bidirectional KD operations, which is defined as
\begin{equation}\label{eq:10}
\begin{aligned}
\noindent L_{dcm_2}(\hat P_{t}, \hat P_{s}) = \sum\limits_{\{(i,j) | 1\leq {i,j} \leq K+1, i \neq j\}}^{} L_{kd}(\hat P_{t_i}, \hat P_{s_j}). \\
\end{aligned}
\end{equation}

\begin{algorithm}[htb]
\SetAlgoNoLine
\SetKwInOut{Input}{Input}
\SetKwInOut{Output}{Output}
\Input{Training data $\{{X,Y}\}$, two CNN models $W_t$ and $W_s$, classifier locations ${\{(t_k,s_k)\}}_{k=1}^{K+1}$, learning rate $\gamma_i$}
Initialise $W_t$ and $W_s$, $i=0$\;
\Repeat{Converge}{
    $i \leftarrow i+1$, update $\gamma_i$\;
    \textbf{1.} Randomly sample a batch of data from $\{{X,Y}\}$\;
    \textbf{2.} Compute knowledge set ${\{(\hat P_{t_k},\hat P_{s_k})\}}_{k=1}^{K+1}$ at all supervised layers of two models by Eq.~\ref{eq:03}\;
    \textbf{3.} Compute loss $L_t$ and $L_s$ by ~Eq.~\ref{eq:07}, ~Eq.~\ref{eq:08}, ~Eq.~\ref{eq:09}, and ~Eq.~\ref{eq:10} \;
    \textbf{4.} Calculate gradients and update parameters:\\ $~~~W_t \leftarrow W_t - \gamma_i \frac{\partial L_t}{\partial W_t}$, $W_s \leftarrow W_s - \gamma_i \frac{\partial L_s}{\partial W_s}$
}
\caption{The DCM algorithm}
\label{alg:dcm}
\vskip -0.2 in
\end{algorithm}

In~Eq.~\ref{eq:09} and~Eq.~\ref{eq:10}, $L_{kd}$ is computed with the modified cross-entropy loss defined by~Eq.~\ref{eq:02}. It matches the probabilistic outputs from any pair of the supervised layers in the teacher and student networks. According to the above definitions, it can be seen: bidirectional KD operations are performed not only between the same-staged supervised layers but also between the different-staged supervised layers of the teacher and student networks. \textit{Benefiting from the well-designed auxiliary classifiers, such two types of cross-layer bidirectional KD operations are complimentary to each other as validated in the experiments}. Enabling dense cross-layer bidirectional KD operations resembles a dynamic knowledge synergy process between two networks for the same task. The training algorithm of our DCM is summarized in Algorithm~\ref{alg:dcm}.

\textbf{Connections to DML and KD}. Regardless of the selection of measure function (Eq.~\ref{eq:02} or Eq.~\ref{eq:05}) for matching probabilistic outputs, in the case where $Q=\emptyset$ meaning the supervision is only added to the last layer of the teacher and student networks, DCM becomes DML. In the extreme case where $Q=\emptyset$ and $L_{t}$ is frozen, DCM becomes KD. Therefore, DML and KD are two special cases of DCM. Like KD and DML, DCM can be easily extended to handle more complex training scenarios where there are more than two neural networks. We leave this part as future research once a distributed system is available for training.

\textbf{Setting of $\bm{Q}$}. In DCM, forming cross-layer bidirectional KD pairs to be connected depends on how to set $Q$. Setting $Q$ needs to consider two basic questions: (1) Where to place auxiliary classifiers? (2) How to design their structures? Modern CNNs adopt a similar hierarchical structure consisting of several stages having different numbers of building blocks, where each stage has a down-sampling layer. In light of this, \textit{to the first question, we use a practical principle, adding auxiliary classifiers merely to down-sampling layers of a backbone network}~\cite{ref44,ref41,ref54,ref58}. Existing works~\cite{ref44,ref58} showed that simple auxiliary classifiers usually worsen the training of modern CNNs as they have no convergence issues. Inspired by them, \textit{to the second question, we use a heuristic principle, making the paths from the input to all auxiliary classifiers have the same number of down-sampling layers as the backbone network, and using backbone's building blocks to construct auxiliary classifiers with different numbers of building blocks and convolutional filters}. Finally in DCM, we enable dense two-way KDs between all layers added with auxiliary classifiers. Although the aforementioned setting may not be the best, it enjoys easy implementation and satisfied results on many CNNs as validated in our experiments.

\section{Experiments}
In this section, we describe the experiments conducted to evaluate the performance of DCM. We first compare DCM with DML~\cite{ref20} which is closely related to our method. We then provide more comprehensive comparisons for a deep analysis of DCM. All methods are implemented with PyTorch~\cite{ref47}. For fair comparisons, the experiments with all methods are performed under the exactly same settings for the data pre-processing method, the batch size, the number of training epochs, the learning rate schedule, and the other hyper-parameters.

\subsection{Experiments on CIFAR-100}
First, we perform experiments on the CIFAR-100 dataset with a variety of knowledge transfer settings.

%\noindent{
{\textbf{CIFAR-100 dataset}. It contains 50000 training samples and 10000 test samples, where samples are $32\times32$ color images collected from 100 object classes~\cite{ref45}. We use the same data pre-processing method as in~\cite{ref06,ref20}. For training, images are padded with 4 pixels to both sides, and $32\times32$ crops are randomly sampled from the padded images and their horizontal flips, which are finally normalized with the per-channel mean and std values. For evaluation, the original-sized test images are used.

%\noindent
{\textbf{Implementation details.} We consider 4 state-of-the-art CNN architectures including: (1) ResNets~\cite{ref06} with depth 110/164; (2) DenseNet~\cite{ref08} with depth 40 and growth rate 12; (3) WRNs~\cite{ref07} with depth 28 and widening factor 4/10; (4) MobileNet~\cite{ref50} as used in~\cite{ref20}. We use the code released by the authors to train each CNN backbone. In the experiments, we consider two training scenarios: (1) Two CNNs with the same backbone (e.g., WRN-28-10 \& WRN-28-10); (2) Two CNNs with the different backbones (e.g., WRN-28-10 \& ResNet-110). In the first training scenario, for ResNets, DenseNet and WRNs, we use the same settings as reported in the original papers~\cite{ref06,ref08,ref07}. For MobileNet, we use the same setting as ResNets, following DML~\cite{ref20}. \textit{In the second training scenario, we use the training setting of the network having better capacity to train both networks}. In our method, we append two auxiliary classifiers to the different-staged layers of each CNN backbone. Specifically, we add each auxiliary classifier after the corresponding building block having a down-sampling layer. All auxiliary classifiers have the same building blocks as in the backbone network, a global average pooling layer and a fully connected layer. The differences are the number of building blocks and the number of convolutional filters. \textit{Detailed designs of auxiliary classifiers and training hyper-parameter settings are provided in the supplementary material}. For each joint training case, we run each method 5 times and report ``mean(std)" error rates. All models are trained on a server using 1/2 GPUs according to the GPU memory requirement.

\begin{table*}[t]
\begin{center}
\caption{Result comparison on the CIFAR-100 dataset. WRN-28-10(+) denotes the models trained with dropout. Bolded results show the accuracy margins of DCM compared to DML. \textit{In this paper, for each joint training case on the CIFAR-100 dataset, we run each method 5 times and report ``mean(std)" top-1 error rates (\%). Results of all methods are obtained with the exactly same training hyper-parameters, and our CNN baselines mostly have better accuracies compared to the numbers reported in their original papers~\cite{ref06,ref08,ref07,ref20}}.
}
\vskip -0.05 in
\label{1}
\scalebox{0.65}{
\begin{tabular}{c|c|c|c|c|c|c|c}
\hline
\Xhline{1.5\arrayrulewidth}
\multicolumn{2}{c|}{Networks}&\multicolumn{2}{c|}{Ind(baseline)}&\multicolumn{2}{c|}{DML}&\multicolumn{2}{c}{DCM}\\
\hline
Net1 & Net2 & Net1 & Net2 & Net1 & Net2 & Net1$|$\textbf{DCM-DML} & Net2$|$\textbf{DCM-DML}\\
%\hline
%ResNet-110 & ResNet-110 & 27.66(0.60) & 27.66(0.60) & 25.46(0.52) & 25.30(0.47) & 25.37(0.29) & \textbf{25.18(0.15)}\\
\hline
\Xhline{1.5\arrayrulewidth}
ResNet-164 & ResNet-164 & 22.56(0.20) & 22.56(0.20) & 20.69(0.25) & 20.72(0.14) & 19.57(0.20)$|$\textbf{1.12} & 19.59(0.15)$|$\textbf{1.13}\\
\hline
WRN-28-10 & WRN-28-10 & 18.72(0.24) & 18.72(0.24) & 17.89(0.26) & 17.95(0.07) & 16.61(0.24)$|$\textbf{1.28} & 16.65(0.22)$|$\textbf{1.30}\\
\hline
DenseNet-40-12 & DenseNet-40-12 & 24.91(0.18) & 24.91(0.18)  & 23.18(0.18) & 23.15(0.20) & 22.35(0.12)$|$\textbf{0.83} & 22.41(0.17)$|$\textbf{0.74}\\
%\hline
%MobileNet & MobileNet & 23.60(0.22) & 23.60(0.22) & 21.94(0.16) & 21.90(0.28) & 20.74(0.11)$|$\textbf{1.20} & 20.47(0.04)$|$\textbf{1.43}\\
%MobileNet & MobileNet & 23.60(0.22) & 23.60(0.22) & 21.94(0.16) & 21.90(0.28) & 20.74(0.11) & \textbf{20.47(0.04)}\\
\hline
\Xhline{1.5\arrayrulewidth}
WRN-28-10 & ResNet-110 & 18.72(0.24) & 26.55(0.26) & 17.99(0.24) & 24.42(0.19) & 17.82(0.14)$|$\textbf{0.17} & 22.99(0.30)$|$\textbf{1.43}\\
\hline
WRN-28-10 & WRN-28-4 & 18.72(0.24) &  21.39(0.30) & 17.80(0.11) & 20.21(0.16) & 16.84(0.08)$|$\textbf{0.96} & 18.76(0.14)$|$\textbf{1.45}\\
\hline
WRN-28-10 & MobileNet & 18.72(0.24) & 26.30(0.35) & 17.24(0.13) & 23.91(0.22) & 16.83(0.07)$|$\textbf{0.41} & 21.43(0.20)$|$\textbf{2.48}\\
\hline
\Xhline{1.5\arrayrulewidth}
WRN-28-10(+) & WRN-28-10(+) & 18.64(0.19) & 18.64(0.19) & 17.62(0.12) & 17.61(0.13) & 16.57(0.12)$|$\textbf{1.05} & 16.59(0.15)$|$\textbf{1.02}\\
\hline
\Xhline{1.5\arrayrulewidth}
\end{tabular}
}
\end{center}
\vskip -0.30 in
\end{table*}

\begin{figure}[t]
\begin{center}
\includegraphics[width=0.6\linewidth]{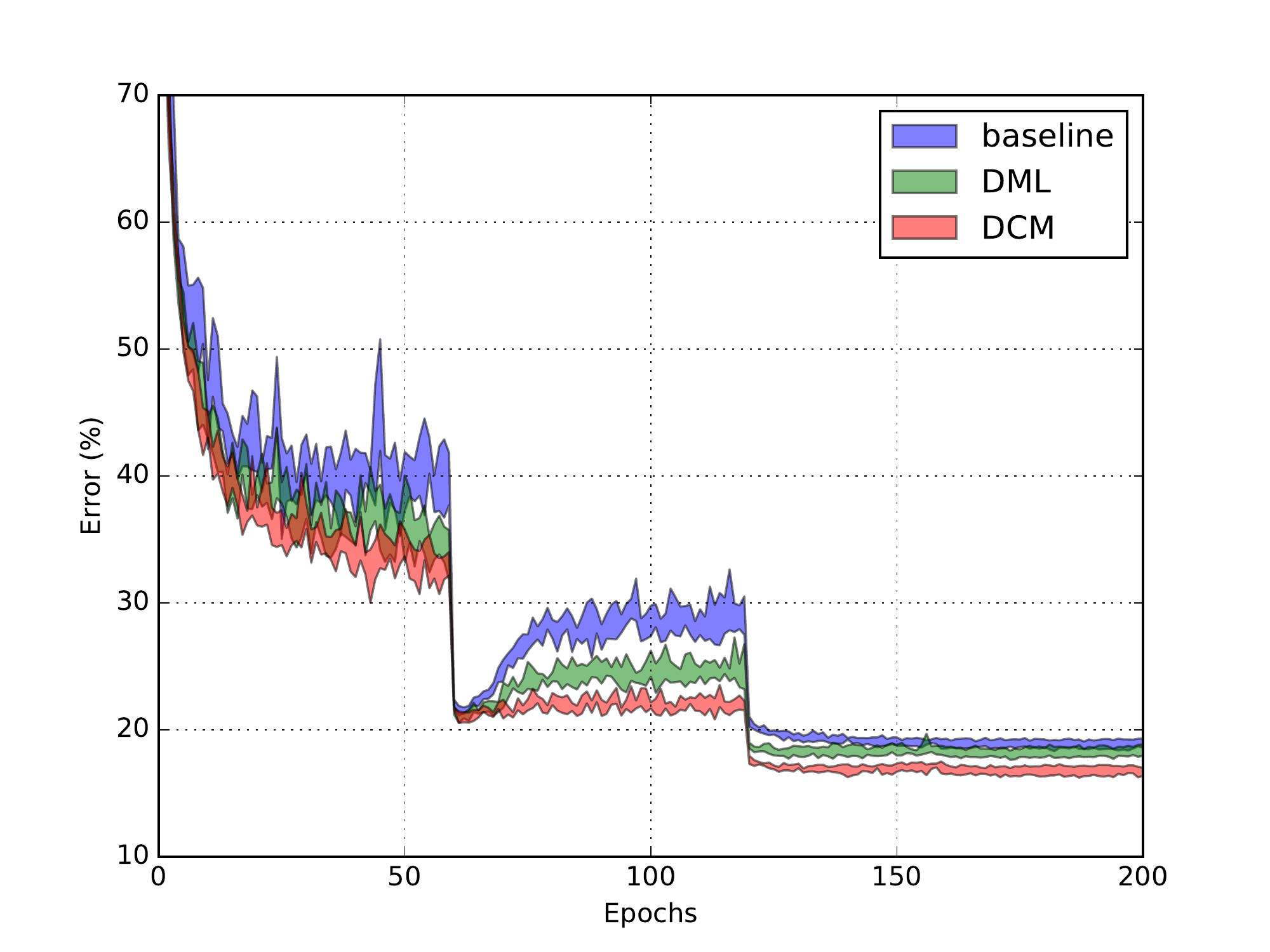}
\end{center}
\vskip -0.20 in
   \caption{Comparison of test curves at the different stages of jointly training two WRN-28-10 models. We show the range over 5 runs. Compared to the independent training method (baseline) and DML, DCM shows stably better performance during the whole training, and finally converges with the best accuracy on the test set.}
\label{fig:error}
\vskip -0.2 in
\end{figure}

\textbf{First training scenario.} Results of training two models with the same backbone are shown in the first part of Table~\ref{1} from which we can find: (1) Both DML and DCM obviously improve the model performance compared to the independent training method; (2) Generally, DCM performs better than DML. Taking the set of models having better mean accuracy as the example, the ResNet-164, WRN-28-10 and DenseNet-40-12 models trained with DCM show $1.12\%$, $1.28\%$ and $0.80\%$ average margins to the models trained with DML respectively; (3) The accuracy gain of DCM against DML shows a trend: the higher the network capacity, the larger the accuracy gain.

\textbf{Second training scenario.} The second part of Table~\ref{1} provides the results of training two models with the different backbones, from which we can make similar observations as in the first training scenario. Besides, we can find another critical observation: Two networks with different capacities have different accuracy improvements. Comparatively, the lower-capacity ResNet-110/MobileNet/WRN-28-4 can benefit more from the high-capacity WRN-28-10 for both DML and DCM, and the corresponding accuracy improvement becomes much more large with DCM. For example, the WRN-28-4/MobileNet model trained with DCM shows $18.76\%/21.43\%$ mean error rate, outperforming the DML counterpart by $1.45\%/2.48\%$ margin.

The aforementioned experiments clearly validate the effectiveness of our method. Fig.~\ref{fig:error} shows an illustrative comparison of test curves at the different stages of training two WRN-28-10 jointly with three different methods.

\subsection{Experiments on ImageNet}
Next, we perform experiments to validate the generalization ability of our method to a much larger dataset.

\textbf{ImageNet classification dataset}. It has about 1.2 million training images and 50 thousand validation images including 1000 object classes~\cite{ref46}. For training, images are resized to $256\times256$, and $224\times224$ crops are randomly sampled from the resized images or their horizontal flips normalized with the per-channel mean and std values. For evaluation, we report top-1 and top-5 error rates using the center crops of resized validation data.

\textbf{Implementation details.} On the ImageNet classification dataset, we use popular ResNet-18/50~\cite{ref06} and MobileNetV2~\cite{ref50-1} as the backbone networks, and consider the two same training scenarios as on the CIFAR-100 dataset. For all these CNN backbones, we use the same settings as reported in the original papers. In our method, we add two auxiliary classifiers to the different-staged layers of each CNN backbone. The auxiliary classifiers are constructed with the same building block as in the backbone network. The differences are the number of building blocks and the number of convolutional filters. \textit{Detailed designs of auxiliary classifiers and training hyper-parameter settings are provided in the supplementary material}. For a concise comparison, we use the conventional data augmentation but not aggressive data augmentation methods. All models are trained on a server using 8 GPUs.

\textbf{Results comparison.} The results are summarized in Table~\ref{2}. It can be seen that both DML and DCM bring noticeable accuracy improvements to the baseline model in the first scenario of training two networks with the same structure jointly, and DCM is better than DML. Comparatively, the better one of two ResNet-18 models trained by DCM shows $28.67\%/9.71\%$ top-1/top-5 error rate which outperforms the baseline model with a margin of $2.41\%/1.46\%$. Impressively, our DCM shows at most $1.04\%/0.76\%$ accuracy improvement to DML on the MobileNetV2 model. These results are consistent with the results of training two networks with the same backbone on the CIFAR-100 dataset. In the second scenario of jointly training two different CNN backbones, the lower-capacity ResNet-18 benefits more from the high-capacity ResNet-50 than the reverse one for both DML and DCM, and the corresponding accuracy improvement becomes much larger by using DCM. Specifically, the ResNet-18 model trained by DCM can even reach $27.93\%/9.19\%$ top-1/top-5 error rate, showing $3.15\%/1.98\%$ and $0.72\%/0.3\%$ gain to the model trained with the independent training method and DML respectively. Although we use the conventional data augmentation, the best ResNet-18 model trained with our DCM shows $2.5\%$ top-1 accuracy gain against the model (trained with aggressive data augmentation methods) released at the official GitHub page of Facebook~\footnote{\url{https://github.com/facebook/fb.resnet.torch}}.

\begin{table*}[t]
  \begin{center}
  \caption{Result comparison on the ImageNet classification dataset. For each network, we report top-1/top-5 error rate (\%). Bolded results show the accuracy margins of DCM compared to the independent training method/DML.}
  \vskip -0.15 in
  \label{2}
  \scalebox{0.55}{
    \begin{tabular}{c|c|c|c|c|c|c|c}
      \hline
      \Xhline{1.5\arrayrulewidth}
      \multicolumn{2}{c|}{Networks}&\multicolumn{2}{c|}{Ind(baseline)}&\multicolumn{2}{c|}{DML}&\multicolumn{2}{c}{DCM}\\
      \hline
      Net1 & Net2 & Net1 & Net2 & Net1 & Net2 & Net1$|$\textbf{DCM-Ind}$|$\textbf{DCM-DML} & Net2$|$\textbf{DCM-Ind}$|$\textbf{DCM-DML}\\
      \hline
      \Xhline{1.5\arrayrulewidth}
      ResNet-18 & ResNet-18 & 31.08/11.17 & 31.08/11.17 & 29.13/9.89 & 29.25/10.00 & 28.67/9.71$|$\textbf{2.41/1.46}$|$\textbf{0.46/0.18} & 28.74/9.74$|$\textbf{2.34/1.43}$|$\textbf{0.51/0.26}\\
      \hline
      MobileNetV2 & MobileNetV2 & 27.80/9.50 & 27.80/9.50 & 26.61/8.85 & 26.78/8.97 & 25.62/8.16$|$\textbf{2.18/1.34}$|$\textbf{0.99/0.69} & 25.74/8.21$|$\textbf{2.06/1.29}$|$\textbf{1.04/0.76}\\
      %\Xhline{1.5\arrayrulewidth}
      \hline
      ResNet-50 & ResNet-18 & 25.47/7.58 & 31.08/11.17 & 25.24/7.56 & 28.65/9.49 & 24.92/7.42$|$\textbf{0.55/0.16}$|$\textbf{0.32/0.14} & 27.93/9.19$|$\textbf{3.15/1.98}$|$\textbf{0.72/0.30}\\
      %\hline
      %ResNet-50 & MobileNetV2 & 25.47/7.58 & 27.80/9.50 & 24.87/7.35 & 26.15/8.64 & 24.32/7.02$|$\textbf{0.85/0.56}$|$\textbf{0.55/0.33} & 25.23/7.97$|$\textbf{2.57/1.53}$|$\textbf{0.92/0.67}\\
      \hline
      \Xhline{1.5\arrayrulewidth}
    \end{tabular}
    }
  \end{center}
  \vskip -0.1 in
\end{table*}

\begin{table}[t]
  \begin{center}
  \caption{Result comparison of jointly training two WRN-28-10 models on the CIFAR-100 dataset using different layer location settings for placing auxiliary classifiers. C1 denotes the default classifier over the last layer of the network, and C2, C3 and C4 denote 3 auxiliary classifiers with the increased layer distance to C1 (see supplementary material for details). We report ``mean(std)" error rates (\%) over 5 runs.}
  \label{3}
  \vskip -0.05 in
  \scalebox{0.75}{
      \begin{tabular}{c|c|c}
      \hline
      \Xhline{1.5\arrayrulewidth}
      \multirow{2}{*}{Classifier locations} & \multicolumn{2}{c}{WRN-28-10}\\\cline{2-3}
      & Net1 & Net2\\\hline
      \Xhline{1.5\arrayrulewidth}
      baseline & 18.72(0.24) & 18.72(0.24)\\\hline
      C1+C4     & 17.16(0.14) & 17.25(0.15)\\
      C1+C3     & 16.89(0.21) & 17.04(0.06)\\
      C1+C2     & 17.40(0.20) & 17.38(0.17)\\
      C1+C2C3(default)   & 16.61(0.24) & 16.65(0.22)\\
      C1+C2C3C4 & \textbf{16.59}(0.12) & 16.73(0.17)\\\hline
      \Xhline{1.5\arrayrulewidth}
    \end{tabular}
    }
  \end{center}
      \vskip -0.20 in
\end{table}

\subsection{Deep Analysis of DCM}
Finally, we conduct extensive ablative experiments on the CIFAR-100 dataset to better understand our method and show its capability to handle more challenging scenarios.

\textbf{Setting of $\bm{Q}$.} Recall that the set $Q$ plays a critical role in DCM. The setting of $Q$ is closely related to two basic questions: (1) Where to place Auxiliary CLassiFiers (ACLFs)? (2) How to design the structure of ACLFs? As we discussed in the method section, we add ACLFs to the down-sampling layers of the network, following the common practices as used in~\cite{ref44,ref41,ref54,ref58}. However, a modern CNN architecture usually has several (e.g., 3/5) down-sampling layers, and thus there exist many layer location combinations for placing ACLFs. To this question, we conduct ablative experiments to jointly train two WRN-28-10 models considering different settings by adding ACLFs to at most three down-sampling layers. The results summarized in Table~\ref{3} show that the 2-ACLF model brings relatively large gain compared to the 1-ACLF model, while the 3-ACLF model gives negligible gain compared to the 2-ACLF model. Therefore, we add 2 ACLFs as the default setting of DCM for a good accuracy-efficiency trade-off. To the second question, we evaluate two additional kinds of ACLFs besides the default ACLFs used in DCM. Results are shown in Table~\ref{4} where ``APFC" refers to a structure that uses an average pooling layer to down-sample input feature maps and then uses a fully connected layer to generate logits. ``Narrow" refers to a narrower version (smaller width multipliers) compared to the default ACLF design. It can be seen that simple ACLFs may hurt performance sometimes (similar experiments are also provided in~\cite{ref44,ref41,ref58}), in which scenario our method has a small gain. Comparatively, large accuracy gains are obtained in the other two cases, therefore relatively strong ACLFs are required to make our method work properly. After training, ACLFs are discarded, and thus there are no extra parameters added to final models.

\begin{table}[t]
    \centering
    \caption{Result comparison of jointly training DenseNet-40-12 and WRN-28-10 on the CIFAR-100 dataset using different types of auxiliary classifiers. We report ``mean(std)" error rates (\%) over 5 runs.}
    \vskip -0.05 in
    \label{4}
\scalebox{0.75}{
\begin{tabular}{c|c|c}
\hline
\Xhline{1.5\arrayrulewidth}
 Net1/Net2 & Classifier type & DCM\\\hline
\Xhline{1.5\arrayrulewidth}
\multirow{2}{*}{DenseNet-40-12} %& baseline & 24.91(0.18) & 24.91(0.18)\\
                   & APFC  & 25.10(0.25)\\
                   & Narrow  & 22.45(0.25)\\
                   & default & \textbf{22.35}(0.12)\\\hline
\multirow{2}{*}{WRN-28-10} %& baseline & 18.72(0.24) & 18.72(0.24)\\
                   & APFC  & 18.23(0.10)\\
                   & Narrow & 16.88(0.17)\\
                   & default & \textbf{16.61}(0.24)\\
\hline
\Xhline{1.5\arrayrulewidth}
\end{tabular}}
 \vskip -0.0 in
\end{table}

\begin{table}[t]
\begin{center}
\caption{Result comparison of training two DenseNet-40-12 models jointly on the CIFAR-100 dataset using different settings of cross-layer bidirectional KD. DCM-1/DCM-2 performs KD operations between the same-staged/different-staged layers. We report ``mean(std)" error rates (\%) over 5 runs.}
\vskip -0.05 in
\label{5}
\scalebox{0.75}{
\begin{tabular}{c|c|c}
\hline
\Xhline{1.5\arrayrulewidth}
\multirow{2}{*}{Method} & \multicolumn{2}{c}{Error (\%)}\\
\cline{2-3}
& Net1 & Net2\\
\hline
\Xhline{1.5\arrayrulewidth}
baseline & 24.91(0.18) & 24.91(0.18)\\
\hline
DML & 23.18(0.18) & 23.15(0.20)\\
\hline
DML + DS & 23.18(0.33) & 23.08(0.28)\\
\hline
DCM-1 & 22.86(0.16) & 22.89(0.14)\\
\hline
DCM-2 & 22.43(0.25) & 22.51(0.18)\\
%\hline
%DCM (top-down) & 23.08(0.15) & 23.03(0.12)\\
%\hline
%DCM (bottom-up) & 23.88(0.13) & 23.98(0.20)\\
\hline
DCM & \textbf{22.35(0.12)} & \textbf{22.41(0.17)}\\
\hline
\Xhline{1.5\arrayrulewidth}
\end{tabular}
}
\end{center}
\vskip -0.20 in
\end{table}

\textbf{Analysis of cross-layer bidirectional KDs.}
Recall that our DCM presents two cross-layer bidirectional KD designs (between either the same-staged or different-staged layers) to mutually transfer knowledge between two networks. In order to study their effects, we conduct two experiments in which we keep either the first or second bidirectional KD design. In the experiments, we consider the case of jointly training two DenseNet-40-12 models. Surprisingly, the results provided in Table~\ref{5} show that the second design brings larger performance gain than the first design, which means knowledge transfer between the different-staged layers is more effective. Because of the introduction of well-designed auxiliary classifiers, DCM enables much more diverse and effective bidirectional knowledge transfer which improves joint training performance considerably.

\textbf{Accuracy gain from deep supervision.}
There is a critical question to DCM: Is the performance margin between DCM and DML mostly owing to the auxiliary classifiers added to certain hidden layers of two networks as they have additional parameters? To examine this question, we conduct extensive experiments considering two different settings: (1) For DCM, we remove all the bidirectional KD connections except the one between the last layer of two DenseNet-40-12 backbones while retaining auxiliary classifiers. This configuration can be regarded as a straightforward combination of DML and Deep Supervision (DS); (2) Further, we add auxiliary classifiers to individual CNN backbones, and train each of them with DS independently while train two same backbones with DCM simultaneously. The results under the first setting are provided in Table~\ref{5}, denoted as DML + DS. It can be observed that the combination of DML and DS only brings $0.07\%$ average improvement to DenseNet-40-12, which only occupies $8.75\%$ of the total margin brought by DCM. The results under the second setting are summarized in Table~\ref{6}. It can be noticed that the average gain of DS to each baseline model is less than $0.85\%$ in the most cases. Comparatively, DCM shows consistently large accuracy improvements over the baseline models, ranging from $2.07\%$ to $2.99\%$.

\begin{table}[t]
  \centering
\caption{Comparison of Deep Supervision (DS) and DCM on the CIFAR-100 dataset. We report ``mean(std)" error rates (\%) over 5 runs.}
\vskip -0.05 in
\label{6}
\scalebox{.75}{
\begin{tabular}{c|c|c|c}
\hline
\Xhline{1.5\arrayrulewidth}
Network & baseline &DS & DCM\\
\hline
\Xhline{1.5\arrayrulewidth}
ResNet-164     & 22.56(0.20) & 21.38(0.32) & \textbf{19.57(0.20)}\\\hline
%ResNet-110     & 27.66(0.60) & 26.95(0.51) & \textbf{25.18(0.15)}\\\hline
DenseNet-40-12 & 24.91(0.18) & 24.46(0.22) & \textbf{22.35(0.12)}\\\hline
WRN-28-10      & 18.72(0.24) & 18.32(0.13) & \textbf{16.61(0.24)}\\\hline
WRN-28-10(+)   & 18.64(0.19) & 17.80(0.29) & \textbf{16.57(0.12)}\\\hline
\Xhline{1.5\arrayrulewidth}
%MobileNet      & 23.60(0.22) & 22.98(0.17) & \textbf{20.47(0.04)}\\\hline
\end{tabular}}
\vskip -0.20 in
\end{table}

\textbf{Comparison with KD and its variants}.
\textit{Note that a fair comparison of DCM/DML with KD and its variants is impractical as the training paradigm is quite different}. Here we illustratively study how much KD will work in our case, via using a pre-trained WRN-28-10 to guide the training of a ResNet-110. Surprisingly, KD shows a slightly worse mean error rate than baseline ($26.66\%$ vs. $26.55\%$). We noticed that during training, the soft labels generated by the teacher (WRN-28-10) are not so ``soft'' and the accuracy of these soft labels is very high ($\sim 99\%$ on the CIFAR-100 dataset, meaning the model usually fits training data ``perfectly''). These soft labels don't provide any more useful guidance than the hard labels and cause overfitting somehow. Using DML or DCM, the soft labels are generated dynamically as the teacher and student networks are jointly trained from the scratch, so they are comparatively softer and contain more useful guidance at every training iteration. \textit{Besides, we provide horizontal comparisons of DCM with KD variants in the supplementary material}.

\textbf{With noisy data.} We also explore the capability of our method to handle noisy data. Following~\cite{ref48,ref49}, we use CIFAR-10 dataset and jointly train two DenseNet-40-12 models as a test case. Before training, we randomly sample a fixed ratio of training data and replace their ground truth labels with randomly generated wrong labels. After training, we evaluate the models on the raw testing set. The results are summarized in Table~\ref{7}. Three corruption ratios 0.2, 0.5, and 0.8 are considered. Compared to the case with 0.2 corruption ratio, the margin between DCM and baseline increases as the corruption ratio increases. One possible explanation of this phenomenon is that DCM behaves as a regularizer. When the training labels get corrupted, the baseline model will try to fit the training data and capture the wrong information, which causes severe overfitting. In the DCM configuration, things are different. Beyond the corrupted labels, the classifiers also get supervision from the soft labels generated by other different-staged or same-staged classifiers. These soft labels can prevent the classifiers from fitting the corrupted data and finally improve the generalization to a certain degree. In normal training without corrupted data, this can also happen. For example, if there is an image of a person with his or her dog, the human-annotated ground truth will be a 1-class label, either ``dog'' or ``person'', but not both. This kind of images can be seen as ``noisy'' data, and this is where soft-labels dynamically generated will kick in.

\textbf{With strong regularization.}
The aforementioned experiments show DCM behaves as a strong regularizer which can improve the generalization of the models. In order to study the performance of DCM under the existence of other strong regularizations, we follow the dropout experiments in~\cite{ref07}. We add a dropout layer with $p=0.3$ after the first layer of every building block of WRN-28-10. The results are shown in Table~\ref{1} as WRN-28-10(+). It can be seen that combining DCM with dropout achieves better performance than DCM, which means DCM is compatible with traditional regularization techniques like dropout.

\textbf{Comparison of efficiency.}
In average, DCM is about $1.5\times$ slower than DML during the training phase due to the use of auxiliary classifiers. However, all auxiliary classifiers are discarded after training, so there is no extra computational cost to the resulting model during the inference phase compared with the independent training method and DML. With DCM, the lower-capacity model has similar accuracy but requires much less computational cost compared to high-capacity model. For example, as shown in Table~\ref{1}, WRN-28-4 models trained with DCM show a mean error rate of 18.76\% which is almost the same to that of WRN-28-10 models trained with the independent training method.

\begin{table}[t]
\begin{center}
\caption{Result comparison on the CIFAR-10 dataset with noisy labels. We jointly train two DenseNet-40-12 models, and report ``mean(std)" error rates (\%) over 5 runs.}
\vskip -0.05 in
\label{7}
\scalebox{0.75}{
\begin{tabular}{c|c|c}
\hline
\Xhline{1.5\arrayrulewidth}
Corruption ratio & Method & Error (\%)\\
\hline
\Xhline{1.5\arrayrulewidth}
\multirow{3}{*}{0.2} & baseline & 9.85(0.24)\\
                     & DML & 8.13(0.14)\\
                     & DCM & \textbf{7.11(0.11)}\\
\hline
\multirow{3}{*}{0.5} & baseline & 17.93(0.39)\\
                     & DML & 14.31(0.30)\\
                     & DCM & \textbf{12.08(0.34)}\\
\hline
\multirow{3}{*}{0.8} & baseline & 35.32(0.42)\\
                     & DML & 32.65(0.96)\\
                     & DCM & \textbf{31.26(0.94)}\\
\hline
\Xhline{1.5\arrayrulewidth}
\end{tabular}
}
\end{center}
\vskip -0.20 in
\end{table}

\section{Conclusions}

In this paper, we present DCM, an effective two-way knowledge transfer method for collaboratively training two networks from scratch. It connects knowledge representation learning with deep supervision methodology and introduces dense cross-layer bidirectional KD designs. Experiments on a variety of knowledge transfer tasks validate the effectiveness of our method.

\clearpage
\section*{\LARGE Appendix: Supplementary Material}
\appendix
  \renewcommand{\appendixname}{Appendix~\Alph{section}}

\section{Experimental Settings}
In this section, we provide detailed settings of the experiments conducted on the CIFAR-100 and ImageNet datasets.

\subsection{Experimental Settings on CIFAR-100}
As stated in Section 4.1 of the main paper, we consider two different scenarios on the CIFAR-100 dataset when jointly training two CNNs from scratch: (1) Two CNNs with the same backbone (e.g., WRN-28-10 \& WRN-28-10); (2) Two CNNs with the different backbones (e.g., WRN-28-10 \& ResNet-110). Generally, we follow the same settings as reported in the original papers~\cite{ref06,ref08,ref07,ref20}. Here, we first describe the training hyper-parameters in the first scenario. At the training phase, for ResNet-110/ResNet-164 and MobileNet, we use SGD with momentum, and we set the batch size as 64, the weight decay as 0.0001, the momentum as 0.9 and the number of training epochs as 200. The initial learning rate is 0.1, and it is divided by 10 every 60 epochs. For DenseNet-40-12, we use SGD with Nesterov momentum, and we set the batch size as 64, the weight decay as 0.0001, the momentum as 0.9 and the number of training epochs as 300. The initial learning rate is set to 0.1, and is divided by 10 at 50\% and 75\% of the total number of training epochs. For WRN-28-10/WRN-28-4, we use SGD with momentum, and we set the batch size as 128, the weight decay as 0.0005, the momentum as 0.9 and the number of training epochs as 200. The initial learning rate is set to 0.1, and is divided by 5 at 60, 120 and 160 epochs. In the second scenario of jointly training two CNNs with the different backbones, we use the training hyper-parameters of WRN-28-10 to train both networks.

\subsection{Experimental Settings on ImageNet}
On the ImageNet dataset, we use popular ResNet-18/ResNet-50~\cite{ref06} and MobileNetV2~\cite{ref50-1} as the backbone networks, and consider two training scenarios which are the same as on the CIFAR-100 dataset. For all these CNN backbones, we use the same settings as reported in the original papers. For the experiments with ResNet backbones, two models with either the same depth configuration or the different depth configurations are trained with SGD for 100 epochs. We set the batch size as 256, the weight decay as 0.0001 and the momentum as 0.9. The learning rate starts at 0.1, and is divided by 10 every 30 epochs. For the experiment with MobileNetV2 backbone, two models with the same configuration are trained with SGD for 150 epochs using batch size 256. The momentum is set as 0.9 and the weight decay is set as 4e-5. The learning rate initiates from 0.05 and declines with a cosine function shaped decay strategy approximating to zero.

\begin{figure}[t]
\centering
\includegraphics [width=0.6\textwidth]{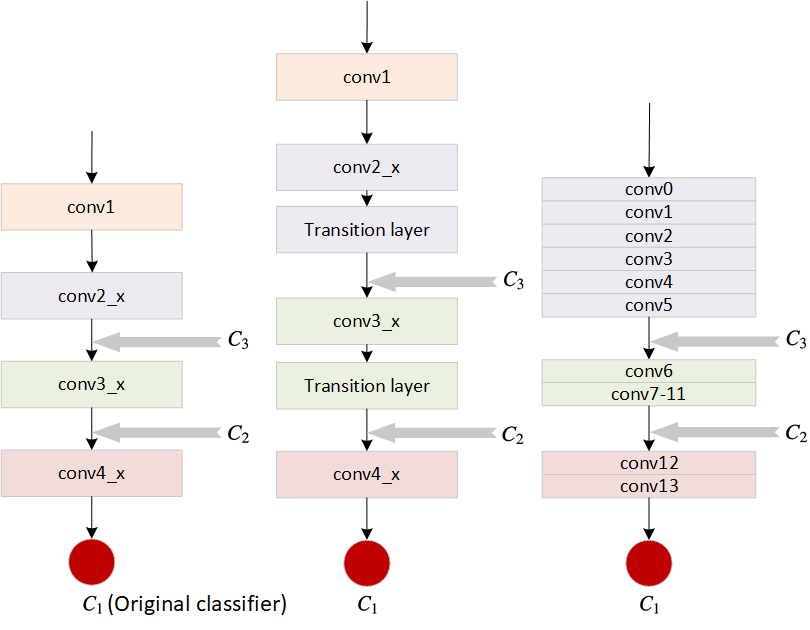}
%\vspace{-3mm}
\caption{Locations of the auxiliary classifiers added to the CNN architectures evaluated on the CIFAR-100 dataset. The left figure is for ResNet-110/ResNet-164 and WRN-28-10/WRN-28-4, and the middle figure is for DenseNet-40-12, and the right figure is for MobileNet. The grey thick arrows indicate the locations where auxiliary classifiers are added. For a backbone network, we denote the original classifier as $C_1$, and denote two auxiliary classifiers as $C_2$ and $C_3$ respectively.}
%\vspace{-3mm}
\label{fig:locations_cifar}
\end{figure}

\section{Structures of Auxiliary Classifiers}
On the CIFAR-100 dataset, we test several kinds of prevailing CNN architectures including ResNet-110/ResNet-164~\cite{ref06}, DenseNet-40-12~\cite{ref08}, WRN-28-10/WRN-28-4~\cite{ref07} and MobileNet~\cite{ref50}. As stated in the main paper, we append well-designed auxiliary classifiers on top of certain different-staged down-sampling layers of a CNN backbone when applying our method to CIFAR-100 and ImageNet datasets. In this section, we provide the structures of our auxiliary classifiers.
\subsection{Auxiliary Classifiers for CIFAR-100}
In this sub-section, we describe the auxiliary classifiers used in the CIFAR-100 experiments.

%\vspace{-3mm}
\begin{table*}[]
\centering
\caption{Details of the convolutional blocks of the auxiliary classifiers added to the ResNet backbones evaluated on the CIFAR-100 dataset. In the table, every cell shows the number of building blocks and the corresponding number of output channels.}
\label{tab:CIFAR_structures_resnet}
\scalebox{0.8}{
\begin{tabular}{|c|c|c|c|c|}
\hline
 & \multicolumn{2}{c|}{ResNet-110} & \multicolumn{2}{c|}{ResNet-164}\\
\hline
 & $C_3$ & $C_2$ & $C_3$ & $C_2$ \\
\hline
 conv1 & - & - & - & -\\
\hline
 conv2\_x &  - & - & - & - \\
\hline
 conv3\_x & {$\begin{bmatrix} 3\times3,\,32 \\ 3\times3,\,32 \end{bmatrix} \times 9$} & - & {$\begin{bmatrix} 1\times1,\,128 \\ 3\times3,\,128 \\ 1\times1,\,128 \end{bmatrix} \times 9$} & -\\
\hline
 conv4\_x & {$\begin{bmatrix} 3\times3,\,64 \\ 3\times3,\,64 \end{bmatrix} \times 9$} & {$\begin{bmatrix} 3\times3,\,128 \\ 3\times3,\,128 \end{bmatrix} \times 18$} & {$\begin{bmatrix} 1\times1,\,256 \\ 3\times3,\,256 \\ 1\times1,\,256 \end{bmatrix} \times 9$} & {$\begin{bmatrix} 1\times1,\,512 \\ 3\times3,\,512 \\ 1\times1,\,512 \end{bmatrix} \times 18$} \\
\hline
\end{tabular}
}
\end{table*}

\begin{table*}[]
\centering
\caption{Details of the convolutional blocks of the auxiliary classifiers added to the DenseNet backbone evaluated on the CIFAR-100 dataset. In the table, every cell shows the number of building blocks and the corresponding growth rate.}
\label{tab:CIFAR_structures_densenet}
\scalebox{0.8}{
\begin{tabular}{|c|c|c|}
\hline
 & \multicolumn{2}{c|}{DenseNet-40-12} \\
\hline
 &  $C_3$ & $C_2$ \\
\hline
 conv1 & - & - \\
\hline
 conv2\_x &  - & - \\
\hline
 conv3\_x &  {$\begin{bmatrix} 3\times3,\,12 \end{bmatrix} \times 12$} & - \\
\hline
 conv4\_x &  {$\begin{bmatrix} 3\times3,\,12 \end{bmatrix} \times 6$} & {$\begin{bmatrix} 3\times3,\,36 \end{bmatrix} \times 12$} \\
\hline
\end{tabular}
}
\end{table*}

\paragraph{Locations.}
In the experiments, we add 2 auxiliary classifiers to every backbone network. We denote the original classifier (i.e., the top-most classifier added to the last layer of a backbone network) as $C_1$ and the auxiliary classifiers as $C_2$ and $C_3$ as shown in Fig.~\ref{fig:locations_cifar}.
%\vspace{-3mm}

\paragraph{Structures.}
In each joint training experiment, every auxiliary classifier is composed of the same building block (e.g., residual block in ResNet) as in the backbone network. The differences lie in the numbers and parameter sizes of convolutional layers. As empirically verified in~\cite{ref44,ref41,ref58}, early layers lack coarse-level features which are helpful for image-level classification. In order to address this problem, we use a heuristic principle making the paths from the input to all classifiers have the same number of down-sampling layers. We detail the hyper-parameter settings of the convolutional layers of auxiliary classifiers w.r.t. different backbone networks in Table~\ref{tab:CIFAR_structures_resnet}, Table~\ref{tab:CIFAR_structures_densenet}, Table~\ref{tab:CIFAR_structures_wrn} and Table~\ref{tab:CIFAR_structures_mobilenet} respectively.

\subsection{Additional Auxiliary Classifiers regarding Experiments for Analyzing How to Set $\bm{Q}$.}
As stated in Section 4.3 of the main paper, in order to analyze how to set $\bm{Q}$ in our method, we first conduct ablative experiments to jointly train two WRN-28-10 models considering different settings by adding auxiliary classifiers to at most three down-sampling layers. Besides two basic auxiliary classifiers $C_2$ and $C_3$ used in our DCM, one additional auxiliary classifier $C_4$ is added after ``conv1'' layer of WRN-28-10 as illustrated in Fig.~\ref{fig:locations_cifar}, and its hyper-parameter setting of the convolutional layers is provided in Table~\ref{tab:CIFAR_structures_wrn}.

As for two additional kinds of auxiliary classifiers evaluated by DCM, namely ``APFC" and ``Narrow", their structures are: (1) ``APFC" is composed of an average pooling layer (with the spatial output size of $4\times4$), a fully connected layer (with the output size of 100) and a softmax function; (2) ``APFC" refers to narrower versions whose growth rate (for DenseNet-40-12) or width (for WRN-28-10) values are at half of our basic designs shown in Table~\ref{tab:CIFAR_structures_densenet} and Table~\ref{tab:CIFAR_structures_wrn} respectively.

\begin{table*}[]
\centering
\caption{Details of the convolutional blocks of the auxiliary classifiers added to the WRN backbones evaluated on the CIFAR-100 dataset. In the table, every cell shows the number of building blocks and the corresponding number of output channels. $C_4$ is used as an additional auxiliary classifier for the analysis of how to set $\bm{Q}$.}
\label{tab:CIFAR_structures_wrn}
\scalebox{0.8}{
\begin{tabular}{|c|c|c|c|c|c|}
\hline
 & \multicolumn{2}{c|}{WRN-28-4} & \multicolumn{3}{c|}{WRN-28-10} \\
\hline
 & $C_3$ & $C_2$ & $C_4 $ & $C_3$ & $C_2$\\
\hline
 conv1 &  - & - & - & - & -\\
\hline
 conv2\_x &  - & - & $\begin{bmatrix} 3\times3,\,160 \\ 3\times3,\,160 \end{bmatrix} \times 4$ & - & - \\
\hline
 conv3\_x & {$\begin{bmatrix} 3\times3,\,128 \\ 3\times3,\,128 \end{bmatrix} \times 4$} & - & {$\begin{bmatrix} 3\times3,\,320 \\ 3\times3,\,320 \end{bmatrix} \times 2$} & {$\begin{bmatrix} 3\times3,\,320 \\ 3\times3,\,320 \end{bmatrix} \times 4$} & - \\
\hline
 conv4\_x & {$\begin{bmatrix} 3\times3,\,256 \\ 3\times3,\,256 \end{bmatrix} \times 2$} & {$\begin{bmatrix} 3\times3,\,512 \\ 3\times3,\,512 \end{bmatrix} \times 4$} &  {$\begin{bmatrix} 3\times3,\,640 \\ 3\times3,\,640 \end{bmatrix} \times 2$} & {$\begin{bmatrix} 3\times3,\,640 \\ 3\times3,\,640 \end{bmatrix} \times 2$} & {$\begin{bmatrix} 3\times3,\,1280 \\ 3\times3,\,1280 \end{bmatrix} \times 4$} \\
\hline
\end{tabular}
}
\end{table*}

\begin{table*}[]
\centering
\caption{Details of the convolutional blocks of the auxiliary classifiers added to the MobileNet backbone evaluated on the CIFAR-100 dataset. In the table, every cell shows the number of convolutional blocks and the number of output channels, and $s2$ denotes the stride of the convolution operation in this layer is 2. Each convolutional block is composed of a $3\times3$ depthwise convolution and a $1\times1$ pointwise convolution. Please see Fig.~\ref{fig:locations_cifar} for different layer locations.}
\label{tab:CIFAR_structures_mobilenet}
\scalebox{0.75}{
\begin{tabular}{|c|c|c|c|c|c|c|c|c|c|c|c|c|}
\hline
 & conv0 & conv1 & conv2 & conv3 & conv4 & conv5 & conv6 & conv7-11 & conv12 &conv13\\
%\hline
%$C_1$ & 32 & 64 & 128 & 128 & 256 & 256 & (512,s2) & 512 $\times$ 5 & (1024,s2), 1024\\
\hline
%$C_3$ & -  & -  & -   & -   & -   & -   & (512,s2){$\begin{bmatrix} 3\times3,\,256 \\ 1\times1,\,512 \end{bmatrix} \times 1$} & 512 $\times$ 3 & (1024,s2), 1024\\
$C_3$ & -  & -  & -   & -   & -   & -   & {$\begin{bmatrix} 3\times3,\,256 s2 \\ 1\times1,\,512 \end{bmatrix}$} & {$\begin{bmatrix} 3\times3,\,512 \\ 1\times1,\,512 \end{bmatrix}$} & {$\begin{bmatrix} 3\times3,\,512 s2 \\ 1\times1,\,1024 \end{bmatrix}$ } & {$\begin{bmatrix} 3\times3,\,1024 \\ 1\times1,\,1024 \end{bmatrix}$}\\
\hline
$C_2$ & -  & -  & -   & -   & -   & -   & -       & -             & {$\begin{bmatrix} 3\times3,\,512 s2 \\ 1\times1,\,2048 \end{bmatrix}$ }  & {$\begin{bmatrix} 3\times3,\,2048 \\ 1\times1,\,2048 \end{bmatrix}$}\\
\hline
\end{tabular}
}
\end{table*}
\subsection{Auxiliary Classifiers for ImageNet}
On the ImageNet dataset, we use popular ResNet-18/ResNet-50 and MobileNetV2 as the backbone networks. In this sub-section, we describe their respective auxiliary classifiers.

\paragraph{Locations.}
The locations of the auxiliary classifiers for ResNet-18/ResNet-50 and MobileNetV2 are shown in Fig.~\ref{fig:locations_imagenet} respectively.

\begin{figure}[t]
\centering
\includegraphics [width=0.5\textwidth]{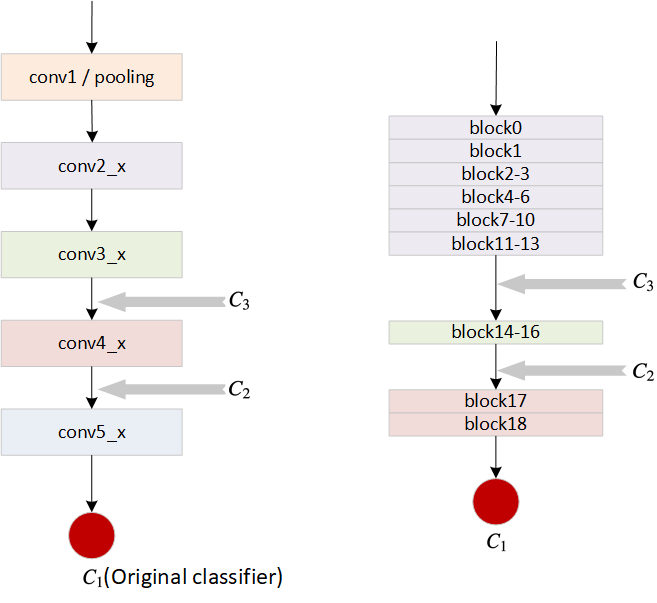}
%\vspace{-3mm}
\caption{Locations of the auxiliary classifiers added to the CNN architectures evaluated on the ImageNet dataset. The left figure is for ResNet-18/ResNet-50, and the right figure is for MobileNetV2. The grey thick arrows indicate the locations where auxiliary classifiers are added. For a backbone network, we denote the original classifier as $C_1$, and two auxiliary classifiers as $C_2$ and $C_3$ respectively.}
\label{fig:locations_imagenet}
%\vspace{-3mm}
\end{figure}

\paragraph{Structures.}
The auxiliary classifiers added to all backbone networks have the same macro-structure. Generally, we design these auxiliary classifiers with the same building blocks as the backbone network. To guarantee that all the paths from the input to different classifier outputs have the same down-sampling process, we design the auxiliary classifiers according to the corresponding building blocks in the backbone network. Taking ResNet-18/ResNet-50 as an example, the auxiliary classifier $C_3$ has its own conv\_4x and conv\_5x acting as down-sampling stages, whose parameter size is smaller than that of the corresponding stages in the backbone network. After these down-sampling stages, there are also a global average pooling layer and a fully connected layer. Auxiliary classifiers for MobileNetV2 are designed with the same principles. We show the details of the convolutional blocks of the auxiliary classifiers for these backbone networks in Table~\ref{tab:resnet structures_imagenet} and Table~\ref{tab:mobilenetv2 structures_imagenet} respectively.

\section{Advantages over KD and Its Variants}

Recall that in this paper we focus on improving two-way knowledge transfer design. Compared with conventional one-way KD and it variants (mostly relying on hidden layer feature/attention distillation), our method has new properties: (1) collaborative training of two models from scratch (no need of the pre-trained and fixed teacher); (2) bidirectional cross-layer knowledge transfer (a smaller model also improves a large model); (3) soft hidden layer knowledge obtained in a supervised manner (no need of the layer-wise feature/attention matching usually conducted with multi-step strategies due to different hidden layer map dimensions between two models). Note that a direct performance comparison of two-way DCM with one-way KD based methods is not fair due to their quite different training paradigms. Recently, a comprehensive benchmark of fourteen state of the art KD based methods on the CIFAR-100 dataset was published in~\cite{ref50-2} from which we can observe: Merely considering the accuracy gain to the smaller student model regardless of training paradigm differences, our results reported in the main paper are mostly better than those one-way KD based methods using the fixed teacher model.

\begin{table*}[]
\centering
\caption{Details of the convolutional blocks of the auxiliary classifiers added to the ResNet backbones evaluated on the ImageNet dataset. In the table, every cell shows the corresponding number of convolutional blocks (including basic blocks for ResNet-18 and bottleneck blocks for ResNet-50) and their parameter sizes. For comparison with the backbone networks, please refer to the Table 1 of the ResNet paper~\cite{ref06}.}
\label{tab:resnet structures_imagenet}
\scalebox{0.8}{
\begin{tabular}{|c|c|c|c|c|}
\hline
 & \multicolumn{2}{c|}{ResNet-18} & \multicolumn{2}{c|}{ResNet-50} \\
 \hline
 & $C_3$ & $C_2$ & $C_3$ & $C_2$ \\
\hline
conv1 & -      & -     & -     & -      \\
\hline
conv2\_x & -      & -     & -     & -      \\
\hline
conv3\_x & -      & -     & -     & -      \\
\hline
conv4\_x & $\begin{bmatrix} 3\times3,\,256 \\ 3\times3,\,256 \end{bmatrix} \times 1$ & -     & $\begin{bmatrix} 1\times1,\,256 \\ 3\times3,\,256 \\ 1\times1,\,1024 \end{bmatrix} \times 3$ & - \\
\hline
conv5\_x & $\begin{bmatrix} 3\times3,\,512 \\ 3\times3,\,512 \end{bmatrix} \times 2$ & $\begin{bmatrix} 3\times3,\,1024 \\ 3\times3,\,1024 \end{bmatrix} \times 2$  & $\begin{bmatrix} 1\times1,\,512 \\ 3\times3,\,512 \\ 1\times1,\,2048 \end{bmatrix} \times 2$ & $\begin{bmatrix} 1\times1,\,1024 \\ 3\times3,\,1024 \\ 1\times1,\,4096 \end{bmatrix} \times 3$ \\
\hline
\end{tabular}
}
%\vskip -0.10 in
\end{table*}
\begin{table*}[]
\centering
\caption{Details of the separable convolutional blocks of the auxiliary classifiers added to the MobileNetV2 backbone evaluated on the ImageNet dataset. In the table, every cell shows the parameter configuration of separable convolutional blocks, and (t,c,n,s) denotes the expansion factor, the number of output channels, the repeated times of bottleneck unit and the stride respectively. For comparison with the backbone network, please refer to the Table 2 of the MobileNetV2 paper~\cite{ref50-1}.}
\label{tab:mobilenetv2 structures_imagenet}
\scalebox{0.8}{
\begin{tabular}{|c|c|c|c|c|c|c|c|c|c|c|c|}
\hline
 & block0 & block1 & block2-3 & block4-6 & block7-10 & block11-13 & block14-16 & block17 & block18\\
\hline
$C_3$ & -  & -  & -   & -   & -  & -  & (6,160,3,2) & (3,320,1,1) & (-,1280,1,1)\\
\hline
$C_2$ & -  & -  & -   & -   & -  & -  & - & (6,480,1,1) & (-,1920,1,1)\\
\hline
\end{tabular}
}
\vskip -0.20 in
\end{table*}

\clearpage
\bibliographystyle{splncs04}
\bibliography{egbib}

\end{document}